\documentclass[10pt, a4paper, twoside]{article}
\usepackage{geometry}
\usepackage[utf8]{inputenc}
\usepackage[english]{babel}
\usepackage[runin]{abstract}
\usepackage{titling}
\usepackage{booktabs}
\usepackage{fancyhdr}
\usepackage{helvet}
\usepackage{csquotes}
\usepackage{graphicx}
\usepackage{blindtext}
\usepackage{parskip}
\usepackage{etoolbox}
\usepackage{graphicx}
\usepackage{todonotes}
\usepackage{amsmath}
\usepackage{cleveref}
\usepackage{glossaries}
\usepackage{comment}
\usepackage{subcaption}
\usepackage{multirow}
\newacronym{ml}{ML}{Machine Learning}
\newacronym{ai}{AI}{Artificial Intelligence}
\newacronym{odd}{ODD}{Operational Design Domain}
\newacronym{dnn}{DNN}{Deep Neural Network}
\newacronym{map}{mAP}{mean Average Precision}
\newacronym{miou}{mIoU}{mean Intersection over Union}
\newacronym{gan}{GAN}{Generative Adversarial Network}
\newacronym{vae}{VAE}{Variational Autoencoder}
\newacronym{peft}{PEFT}{Parameter-Efficient Fine-Tuning}
\newacronym{dps}{DPS}{Dynamic Parameter Selection}
\newacronym{ssl}{SSL}{Self-supervised learning}
\newacronym{mae}{MAE}{Masked Autoencoder}
\newacronym{cl}{CL}{Curriculum Learning}
\newacronym{spl}{SPL}{Self-Paced Learning}
\newacronym{per}{PER}{Prioritized Experience Replay}
\newacronym{spcl}{SPCL}{Self-Paced Curriculum Learning}
\newacronym{plf}{PLF}{Progressive Layer Freezing}
\newacronym{clr}{CLR}{Cyclical Learning Rates}
\newacronym{iou}{IoU}{Intersection over Union}
\newacronym{fr}{FR}{Fabrication Object Ratio}
\newacronym{vr}{VR}{Vanishing Object Ratio}
\newacronym{zod}{ZOD}{Zenseact Open Dataset}
\newacronym{acdc}{ACDC}{Adverse Conditions Dataset with Correspondences}
\newacronym{fpn}{FPN}{Feature Pyramid Network}
\newacronym{rpn}{RPN}{Region Proposal Network}
\newacronym{roi}{ROI}{Region of Interest}

\makenoidxglossaries

\graphicspath{ {./images/} }

\date{}

\makeatletter

\def\fulltitle#1{\def\@fulltitle{#1}}
\def\runningtitle#1{\def\@runningtitle{#1}}
\def\runningauthor#1{\def\@runningauthor{#1}}
\def\affiliation#1{\def\@affiliation{#1}}
\def\department#1{\def\@department{#1}}
\def\memoid#1{\def\@memoid{#1}}
\def\theyear#1{\def\@theyear{#1}}
\def\mydate#1{\def\@mydate{#1}}

\runningtitle{How to write an effective report} 
\author{Your name \and John Doe \and Jane Doe} 
\runningauthor{Your name et al.} 
\affiliation{Your University} 
\department{Your Department} 
\memoid{AI Memo 777} 
\theyear{2018} 
\mydate{August 08, 2018} 

\def\displaymydate{\@mydate}
\def\displaytheyear{\@theyear}
\def\displaymemoid{\@memoid}
\def\displaydepartment{\@department}
\def\displayaffiliation{\@affiliation}
\def\displayrunningauthor{\@runningauthor}
\def\displayrunningtitle{\@runningtitle}

\makeatother

\makeatletter
\patchcmd{\@zfancyhead}{\fancy@reset}{\f@nch@reset}{}{}
\patchcmd{\@set@em@up}{\f@ncyolh}{\f@nch@olh}{}{}
\patchcmd{\@set@em@up}{\f@ncyolh}{\f@nch@olh}{}{}
\patchcmd{\@set@em@up}{\f@ncyorh}{\f@nch@orh}{}{}
\makeatother

\fancypagestyle{firstpage}
{
    \fancyhf{}
    \fancyhead{%
        \begin{center}
            \textsc{\displayaffiliation}\\
            \textsc{\displaydepartment}
        \end{center}
        \vspace*{0.5in}
        \displaymemoid
        \hfill
        \displaymydate
    }
    \fancyfoot[L]{\footnotesize \textbf{\textsc{\displayaffiliation} -- \displaytheyear}}
}
\providecommand{\keywords}[1]{\noindent \textbf{Keywords:} #1} 

\abslabeldelim{:} 
\pretitle{\Large \begin{center}}
\posttitle{\\[2ex] \Large \textbf{by}\end{center}}

\title{\textbf{Structuring a Training Strategy to Robustify Perception Models with Realistic Image Augmentations}} 
\runningtitle{Structuring an Augmented Training Strategy} 
\author{Ahmed Hammam,
Bharathwaj Krishnaswami Sreedhar,
Nura Kawa, \\
Tim Patzelt,
and Oliver De Candido} 
\runningauthor{neurocat} 
\affiliation{neurocat} 
\department{Research and Engineering} 
\memoid{} 
\theyear{2024} 
\mydate{August 30, 2024} 

\begin{document}

\begin{titlepage}

\newcommand{\HRule}{\rule{\linewidth}{0.5mm}} 

\centering
\includegraphics[height=8cm]{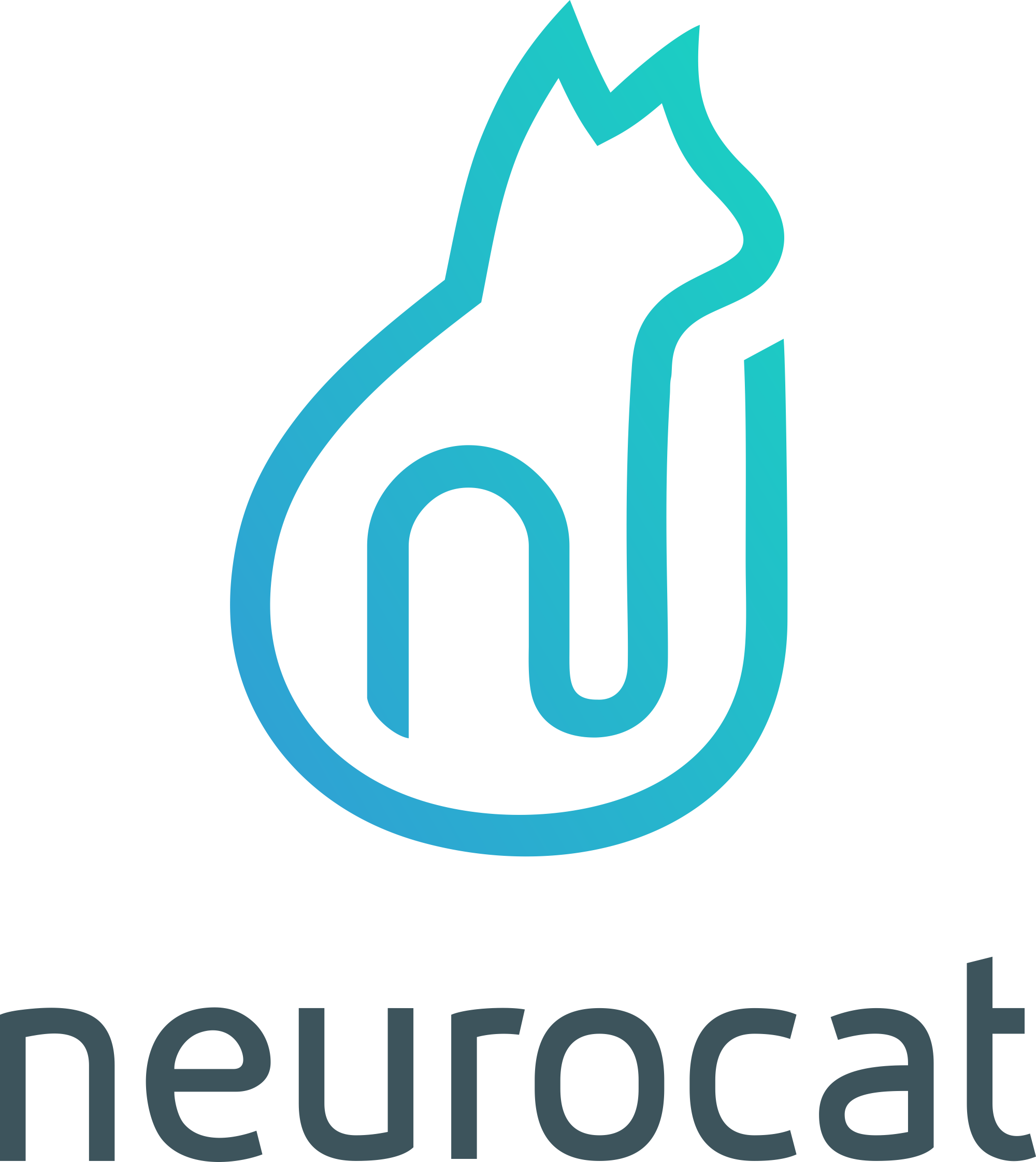}\\[2cm] 
 

\center 


\textsc{\LARGE Technical Report}\\[1.5cm] 

\makeatletter
\HRule \\[0.4cm]
{ \huge \bfseries Structuring a Training Strategy to Robustify Perception Models with Realistic Image Augmentations }\\[0.4cm] 
\HRule \\[1.5cm]
 

\begin{minipage}{0.3\textwidth}
\begin{flushleft} \large

\end{flushleft}
\end{minipage}
~
\begin{minipage}{0.4\textwidth}
\begin{flushright} \large
\end{flushright}
\end{minipage}\\[2cm]
\makeatother


{\large August 30, 2024}\\[2cm] 

\vfill 

\end{titlepage}

\maketitle

\begin{abstract}
    \noindent
    Advancing \gls{ml}-based perception models for autonomous systems necessitates addressing weak spots within the models, particularly in challenging \glspl{odd}. These are environmental operating conditions of an autonomous vehicle which can contain difficult conditions, e.g.,  lens flare at night or objects reflected in a wet street. This report introduces a novel methodology for training with augmentations to enhance model robustness and performance in such conditions. The proposed approach leverages customized physics-based augmentation functions, to generate realistic training data that simulates diverse \gls{odd} scenarios.
    
We present a comprehensive framework that includes identifying weak spots in ML models, selecting suitable augmentations, and devising effective training strategies. The methodology integrates hyperparameter optimization and latent space optimization to fine-tune augmentation parameters, ensuring they maximally improve the ML models’ performance. Experimental results demonstrate improvements in model performance, as measured by commonly used metrics such as \gls{map} and \gls{miou} on open-source object detection and semantic segmentation models and datasets.

Our findings emphasize that optimal training strategies are model- and data-specific and highlight the benefits of integrating augmentations into the training pipeline. By incorporating augmentations, we observe enhanced robustness of \gls{ml}-based perception models, making them more resilient to edge cases encountered in real-world \glspl{odd}. This work underlines the importance of customized augmentations and offers an effective solution for improving the safety and reliability of autonomous driving functions.

The insights presented in this report provide a valuable resource for researchers and practitioners aiming to enhance the robustness of their \gls{ml} models in autonomous systems.

    \keywords{Image Augmentations, Computer Vision, Semantic Segmentation, Object Detection, Training Strategy, Autonomous Vehicles, Operational Design Domain}
\end{abstract}
\glsresetall

\vspace{2.5cm}

\thispagestyle{firstpage}

\pagebreak


\section{Introduction}\label{sec:intro}
In this technical report, we describe our experiences and methodology to find a training strategy to train \glspl{dnn} on augmented images.
Specifically, we investigate \gls{ml}-based perception models for autonomous systems.
Most \gls{ml}-based perception models have weak spots, i.e., data which the model struggles to classify correctly or with high confidence.
These challenges can arise due to the lack of training data in all possible \glspl{odd}.
An \gls{odd} describes the specific domain or the environmental operating conditions where the autonomous vehicle should reliably and safely work, see, e.g.,~\cite{bsi1883}.
Within an \gls{odd}, there might be trigger conditions, e.g., heavy rain, which can lead to errors or faults in the system.
For example, there can be a lack of images in the training dataset with lens flares at night or with objects reflected on wet streets.

Collecting data in all safety-relevant parts of the \gls{odd} comes at a high cost, e.g., a company might have to drive their data collection fleet to an area where fog is more prevalent to collect foggy images.
On top of this, there is a high cost of labeling these data, e.g., for object detection or semantic segmentation tasks.
The cost of labeling a single semantic segmentation image can start at a few dollars and tens of man-hours.
Additionally, one may not have permission to train a \gls{dnn} on data from some specific \glspl{odd}.

One solution to these challenges is to use data augmentations.
These are functions which take an input image, e.g., a driving image taken in good weather, and transform it into an augmented image, e.g., a simple transformation could be a rotation or flipping of the original image.
Image augmentations have been used extensively in training \glspl{dnn} to help them become invariant to these transformations.
For example, we want an object detection model to be able to detect a pedestrian regardless of their posture or walking direction.
However, most augmentation packages, e.g., Albumentations~\cite{albumentations}, offer only simple transformation with limited customizations for specific use-cases.
As such, they are usually used for general performance improvements.

In our work, we take advantage of complex image augmentations which can transform an input image into an image from a specific region of the \gls{odd}.
These image augmentations are highly parameterizable.
They not only use the original image, but also take advantage of segmentation maps and depth maps to create realistic data augmentations.
For example, the depth map can help us wrap the fog augmentation around objects in the scene, and the segmentation map can help us reflect objects on the wet street.
The parameterization of these augmentations allows us to create multiple, different augmented images for each individual input image.
As we will explore, this can help us when finding a training strategy for models to become more robust in specific \glspl{odd}.

In general, when training a \gls{ml} model, we assume that we have a training, validation, and test dataset which come from the same underlying distribution.
That is, we assume that we have enough samples in each dataset of a specific \gls{odd} trigger condition, e.g., images with heavy rain, images with object reflected in wet streets, or images with lens flares, such that we can train on these data.
However, this assumption does not always hold due to the scarcity of these data from specific \glspl{odd}.
Therefore, we have to decide which dataset to add these samples to, i.e., the training, validation, or test dataset.
Our work with complex, parameterizable image augmentations alleviates this challenge, by allowing us to introduce as many images from specific \gls{odd} trigger conditions as we require.
As such, we can use the ground truth images in the validation and test datasets to gain a good estimate of the generalization performance of the trained \gls{dnn}.
We will explore these training strategies more in \Cref{sec:methodology}.

Another approach to training \glspl{dnn} on specific \gls{odd} data is to use synthetic data, e.g., data created using a computer graphics tool like Unreal Engine or data created using a generative model, see, e.g.,~\cite{wu2023ai,marathe2023wedge}.
Using this approach, we can generate unlimited samples in specific \glspl{odd}.
However, it is not clear whether these samples are completely photo-realistic, or if a \gls{dnn} will train on them.
Additionally, it can be difficult when using generative models to control different aspects of the image, e.g., the size of the rain drops or the reflectivity of a wet street.
On the other hand, using image augmentations, we know that the original image is realistic and only need to ensure that the augmentation function generates an augmented image which looks like an image from a specific \gls{odd}, see, e.g.,~\cite{hietala2021closing,Tremblay_2018_CVPR_Workshops,seddik2024bad}.
For aforementioned reasons, we did not explore training a \gls{dnn} on synthetic data instead focusing on augmented data.

Ultimately, we aim to introduce a training strategy allowing us to improve an \gls{ml} models performance on weak spots for a specific \gls{odd} trigger condition.
This technical report is organized as follows: \Cref{sec:related-work} summarizes various approaches to training a model to improve its performance on specific domains.
\Cref{sec:methodology} introduces the details of our proposed methodology to train on augmentations which were specifically designed to reflect real-world \gls{odd} trigger conditions.
\Cref{sec:experiment-setup} and \Cref{sec:experimental-results} discuss the experimental setup and results of various perception models trained with our augmentations, respectively.
Finally, \Cref{sec:conclusion} summarizes our work and discusses future work in this direction.

\section{Literature Review}\label{sec:related-work}

This section examines recent advancements in fine-tuning and generalization methods for \glspl{dnn}, particularly in computer vision. It covers key strategies like data augmentation, synthetic data generation, and fine-tuning, along with \gls{cl}, layer-wise fine-tuning, and hybrid approaches that combine multiple techniques. The goal is to provide a clear overview of how these methods improve the adaptability and performance of pretrained models across different tasks and domains.

\subsection{Data Augmentation and Synthetic Data Generation}
Data augmentation and synthetic data generation are pivotal in enhancing the generalization of \glspl{dnn} for computer vision tasks. Data augmentation involves transforming existing images to increase dataset diversity, with advanced methods like AutoAugment~\cite{cubuk2019autoaugment} and Mixup~\cite{zhang2018mixup} show significant improvements in model performance by creating varied training examples. Synthetic data generation utilizes models such as \glspl{gan} and \glspl{vae} to create artificial data resembling real-world images. These methods are essential for tasks requiring extensive annotated datasets, such as medical imaging \cite{karras2019stylegan} \cite{mumuni2024survey}.

\subsection{Fine-Tuning Pretrained Models}

Fine-tuning pretrained models has been a cornerstone strategy in transferring learned representations from large-scale datasets to more specific tasks, particularly when data is limited. This approach leverages the extensive features learned during pretraining on a broad dataset, such as ImageNet~\cite{deng2009imagenet}, and adapts them to the target task, often leading to significant improvements in performance.

\textbf{\gls{peft}}: As models grow increasingly large, fine-tuning all parameters becomes computationally expensive. \gls{peft} modifies only a subset of parameters while keeping the majority frozen, significantly reducing the computational burden while maintaining performance on the pretraining task and improving performance on the fine-tuning task~\cite{xin2024peft}.

\textbf{\gls{dps}}: Zhang et al. \cite{zhang2022dps} introduced \gls{dps}, which adaptively selects promising sub-networks for updates during fine-tuning based on gradient information. \gls{dps} helps prevent overfitting and maintains stable representations, particularly in low-resource and out-of-domain scenarios.

\textbf{\gls{ssl} and Fine-Tuning}: \gls{ssl} pretrains models on unlabeled data before fine-tuning. Techniques like \glspl{mae} enable models to learn robust features that generalize well across tasks. Sawano et al. \cite{sawano2024ssl} demonstrated the effectiveness of \gls{ssl} in medical imaging, showing how it leads to superior performance in specialized domains.

\textbf{Collateral Damage Mitigation During Fine-Tuning}: Fine-tuning can cause collateral damage, where useful pretraining features are forgotten. Recent strategies focus on selecting pretrain samples likely to be forgotten, thereby mitigating this issue and maintaining a balance between pretrain and fine-tune task performance \cite{collateral2024oxford}.

\subsection{Curriculum Learning and Self-Paced Learning}

\gls{cl} and \gls{spl} are two complementary techniques designed to improve the training efficiency and generalization of \glspl{dnn} by structuring the learning process in a more meaningful and adaptive way.

\gls{cl} involves training a model by gradually increasing the complexity of the examples it encounters, inspired by the way humans learn. Recent research has introduced adaptive curriculum strategies where the difficulty is dynamically adjusted based on the model's performance. For instance, ScreenerNet \cite{meng2023screenernet} predicts the importance of training samples, enabling the main network to learn in an end-to-end self-paced manner, effectively balancing the learning process.

\gls{spl} allows the model to automatically select the order of learning from training data, starting with easier samples and gradually progressing to harder ones. This technique has been extended to various domains, including reinforcement learning and computer vision. For example, the integration of \gls{spl} with \gls{per} in deep reinforcement learning enables dynamic prioritization of experiences, improving convergence and overall performance \cite{jiang2018selfpaced}.

The combination of \gls{cl} and \gls{spl}, often referred to as \gls{spcl}, leverages the strength of both methods. \gls{spcl} adaptively adjusts the learning pace and difficulty level, making it particularly effective in complex tasks such as cross-domain object detection. This hybrid approach has been shown to improve convergence speed and generalization across various domains, including visual recognition and deep reinforcement learning \cite{zhao2023spcl}.

\subsection{Layer-Wise Fine-Tuning}

Layer-wise fine-tuning is an advanced technique in the fine-tuning of pretrained models, particularly effective in scenarios where the new target task is significantly different from the original task. This method involves gradually unfreezing and fine-tuning layers, starting from the higher layers (i.e., layers closer to the output) and moving towards the lower layers (i.e., layers closer to the input). The hierarchical nature of \glspl{dnn} underpins this approach, with lower layers capturing general features useful across tasks, while higher layers capture task-specific features.

\textbf{Fine-Tuning with Frozen Lower Layers}: Kornblith et al. \cite{kornblith2019better} conducted a comprehensive study on different fine-tuning strategies, finding that starting with higher layers and gradually unfreezing lower layers leads to better performance, particularly when there is a significant domain shift between the pretraining and target tasks.

\textbf{Task-Specific Layer-Wise Strategies}: Sun et al. \cite{sun2021task} proposed a dynamic layer-wise fine-tuning strategy that determines which layers to fine-tune based on the complexity and similarity of the target task to the pretraining task. This approach has shown significant improvements in fine-tuning efficiency and model performance.

\textbf{\gls{plf}}: Zhuang et al. \cite{zhuang2020plf} introduced \gls{plf}, where layers are progressively frozen as the model converges, preventing overfitting by preserving well-learned features in the lower layers. \gls{plf} has proven particularly effective in tasks with limited data.

\textbf{Hybrid Approaches Combining Layer-Wise Fine-Tuning with \gls{ssl}}: Chen et al. \cite{chen2020hybrid} explored combining \gls{ssl} pretraining with layer-wise fine-tuning, showing that this hybrid approach significantly improves the effectiveness of fine-tuning in scenarios with scarce labeled data.

\subsection{Learning Rate Scheduling in Fine-Tuning Pretrained Models}

Learning rate scheduling is a crucial technique in fine-tuning pretrained models, directly influencing the model's ability to adapt to new data while preserving learned features. It adjusts the learning rate dynamically during training to optimize performance, balancing rapid adaptation with stability.

Using a high learning rate during fine-tuning can cause unstable updates, disrupting valuable pretraining features, while a low learning rate may slow convergence. Scheduling provides a balanced approach by fine-tuning the learning rate throughout the training process, allowing for precise control over model adaptation.

\textbf{Step Decay}: This method reduces the learning rate at predefined intervals, allowing larger updates early in training and smaller, more refined adjustments later. Smith and Topin \cite{smith2019stepdecay} showed that step decay stabilizes fine-tuning in transfer learning by reducing oscillations in the loss function.

\textbf{Cosine Annealing}: Loshchilov and Hutter \cite{loshchilov2017cosine} introduced cosine annealing, which adjusts the learning rate following a cosine curve. This method helps avoid premature convergence by periodically increasing the learning rate, promoting exploration of new areas in the loss landscape.

\textbf{\gls{clr}}: \gls{clr} \cite{smith2017clr} varies the learning rate cyclically within a specified range, allowing the model to explore different regions of the loss landscape throughout training. Smith \cite{smith2017clr} demonstrated that \gls{clr} enhances fine-tuning performance, especially when there is a domain shift between pretraining and target tasks.

\textbf{Warm Restarts}: Loshchilov and Hutter \cite{loshchilov2017warm} also explored warm restarts, where the learning rate is periodically reset to a higher value before gradually decaying again. This technique helps the model escape local minima and avoid suboptimal solutions during training.

\textbf{Adaptive Learning Rates}: Adam and RMSprop are adaptive learning rate methods that adjust the learning rate based on the gradients' first and second moments. Kingma and Ba \cite{kingma2015adam} introduced Adam, which has become a standard optimizer in fine-tuning due to its ability to handle noisy gradients and sparse data effectively.

\subsection{Regularization Techniques in Fine-Tuning Pretrained Models}

Regularization techniques are essential for enhancing the generalization of \glspl{dnn} during fine-tuning, especially when dealing with small or noisy datasets. These techniques prevent overfitting by penalizing model complexity or introducing randomness during training, ensuring that the model learns generalizable patterns rather than memorizing the training data.

\textbf{Dropout}: Dropout is a widely-used regularization method that randomly sets a fraction of neurons' activations to zero during training, forcing the network to learn redundant representations. Srivastava et al. \cite{srivastava2014dropout} demonstrated that dropout significantly reduces overfitting in large \glspl{dnn}. In fine-tuning, dropout helps the model avoid over-specialization to the new dataset, which is particularly important when this dataset is small or lacks diversity.

\textbf{Weight Decay (L2 Regularization)}: Weight decay, or L2 regularization, penalizes the loss function based on the squared magnitude of the model’s weights. This encourages smaller weights, leading to simpler models less likely to overfit. During fine-tuning, weight decay ensures that the model retains general patterns learned during pretraining while adapting to specific characteristics of the new dataset \cite{loshchilov2017sgdr}.

\textbf{Batch Normalization}: Batch normalization normalizes the inputs to each layer to have zero mean and unit variance, stabilizing the learning process and allowing higher learning rates. It acts as a regularization method by adding a small amount of noise to the inputs of each layer during training. This noise prevents the model from becoming overly dependent on specific examples, enhancing generalization. Batch normalization is particularly useful in fine-tuning as it maintains network stability when adapting to new data \cite{ioffe2015batchnorm}.

\textbf{Data Augmentation as Implicit Regularization}: Data augmentation increases the diversity of training data and also serves as a form of regularization. By presenting varied versions of the training data, such as through rotations or flips, data augmentation forces the model to learn more general features. Zhang et al. \cite{zhang2018mixup} demonstrated the effectiveness of Mixup, a technique blending pairs of training examples, in reducing overfitting and improving generalization during fine-tuning.

\textbf{Stochastic Depth}: Stochastic depth randomly drops entire layers during training, similar to dropout but at the layer level. This encourages the network to learn multiple paths through the network, improving robustness and generalization. Huang et al. \cite{huang2016stochasticdepth} showed that stochastic depth improves training efficiency in very large \glspl{dnn} and enhances performance on unseen data. During fine-tuning, stochastic depth helps prevent overfitting by promoting the learning of more distributed and generalized representations.

\subsection{Hybrid Approaches in Fine-Tuning Pretrained Models}

Hybrid approaches in fine-tuning combine multiple techniques to leverage their strengths, enhancing the generalization and performance of pretrained models on new tasks. These methods typically blend traditional fine-tuning with advanced strategies like \gls{ssl}, data augmentation, or adversarial training, allowing models to better adapt to diverse and challenging datasets.

\textbf{\gls{ssl} Combined with Fine-Tuning}: \gls{ssl} involves pretraining a model on unlabeled data to learn general features before fine-tuning on labeled data. This approach is particularly useful in scenarios where labeled data is scarce but unlabeled data is abundant. For instance, models pretrained using techniques like contrastive learning or \glspl{mae} can be fine-tuned on downstream tasks, resulting in superior performance compared to models trained from scratch. Chen et al. \cite{chen2020ssl} demonstrated that \gls{ssl}-pretrained models, when fine-tuned, achieved state-of-the-art results on various computer vision benchmarks. This approach allows the model to benefit from the broad, general representations learned during \gls{ssl}, which can be finely adjusted to the specific requirements of the target task.

\textbf{Adversarial Training with Fine-Tuning}: Adversarial training involves exposing the model to adversarial examples—inputs intentionally perturbed to cause the model to make incorrect predictions—during fine-tuning. This method strengthens the model's robustness to such perturbations, improving its ability to generalize to new, unseen data. A notable example of this approach is the combination of adversarial training with traditional fine-tuning, which has been shown to enhance the model's resilience against adversarial attacks while maintaining or even improving its performance on standard benchmarks \cite{madry2018adversarial}.

\textbf{Domain Adaptation with Fine-Tuning}: Domain adaptation techniques, when combined with fine-tuning, help models generalize better across different domains. This hybrid approach involves aligning the feature distributions of the source (pretrained) and target (fine-tuning) domains, often through adversarial methods or by minimizing domain discrepancy. Ganin and Lempitsky \cite{ganin2015domain} introduced a method that integrates domain adaptation with fine-tuning, significantly improving the model's performance in cross-domain tasks where the target domain differs substantially from the pretraining domain.

\textbf{Data Augmentation with Fine-Tuning}: Another effective hybrid approach involves integrating sophisticated data augmentation techniques with fine-tuning. Data augmentation can increase the diversity of the training data, exposing the model to a wider variety of scenarios during fine-tuning. For example, Mixup, a data augmentation technique where new samples are generated by blending pairs of existing samples, can be combined with fine-tuning to improve model robustness and generalization. Zhang et al. \cite{zhang2018mixup} showed that Mixup not only reduces overfitting but also improves the performance of models fine-tuned on small or noisy datasets.
The methodology presented in this technical report is motivated by these results.

\section{Methodology – Augmented Training Strategy}\label{sec:methodology}

\subsection{Overview of the Augmented Training Strategy}
\label{sec:overview-augmented-training-strategy}
The augmented training strategy is a systematic approach to enhance the robustness of \glspl{dnn} in the context of computer vision tasks. This process is especially crucial for automated driving systems, where models need to perform reliably across a wide range of environmental conditions, including challenging scenarios and \glspl{odd} trigger conditions, e.g., those with heavy rain, fog, or low visibility. The methodology is composed of several interconnected stages, aimed at (i) identifying model weaknesses; (ii) applying appropriate augmentations to clear weather images; (iii) fine-tuning a \gls{dnn} using those augmented images; and (iv) rigorously evaluating the fine-tuned models to measure the performance and robustness improvement.

\subsubsection{Weak Spot Identification}

The first step in the process is to identify the specific weaknesses of a \gls{dnn}. This involves testing the model across various \gls{odd} scenarios—different environmental and operational conditions that the model might encounter in real-world applications. For example, a model may perform adequately under clear weather conditions but struggle significantly in rainy or foggy environments.

To identify these weaknesses, the model is subjected to a series of robustness tests using a carefully chosen set of metrics. These metrics are task-specific, ensuring that they accurately reflect the model's performance in the context of its intended application. For object detection tasks, common metrics include:

\begin{itemize}
    \item \textbf{\gls{map}:} This metric measures the precision of the model across different object classes, averaging the precision scores to provide an overall performance indicator.
    \item \textbf{Vanishing Ratio:} This ratio measures the occurrence of false negatives, indicating how often the model fails to detect objects that are present in the scene.
    \item \textbf{Fabrication Ratio:} This ratio measures the occurrence of false positives, showing how often the model incorrectly detects objects that are not present.
\end{itemize}

For semantic segmentation tasks, the \textbf{\gls{miou}} is commonly used, which calculates the overlap between the predicted segmentation map and the ground truth segmentation map across different classes.

These metrics help us to pinpoint the scenarios where the model's performance deteriorates or is insufficient, guiding the next steps in the augmented training process.
\Cref{sec:eval-metrics} introduces these metrics in more detail.

\subsubsection{Augmentation Selection and Fine-Tuning}
\label{sec:fine-tuning-augmentations}

Once the weak spots of the model have been identified, the next step is to select and apply appropriate augmentations to the training data. Data augmentation is a technique used to artificially expand the training dataset by applying various transformations to the existing data. 
In our case, these transformations are complex, physics-based functions replicating weather conditions which simulate the challenging conditions where the model struggles, e.g., rain, fog, or snow.

For instance, if the model performs poorly under rainy conditions, rain-specific augmentations are chosen. These augmentations can be parameterized to closely mimic real-world rain scenarios. The parameters might include, e.g., the intensity of the rain, the reflectivity of wet surfaces, or the amount of blur caused by raindrops. The goal is to create a set of augmented images that represent the challenging scenarios as accurately as possible, providing the model with more relevant training data.
Additionally, due to the parameterization of the augmentation functions, we can create multiple, unique augmented images for each input image.
This can be leveraged as part of the augmented training strategy.

\textbf{Fine-Tuning Augmentations}

After selecting suitable augmentations, the next stage is fine-tuning these augmentations to maximize their impact on the model’s performance. This stage is crucial because not all augmentations will have the same effect on the model; some might improve performance significantly, while others may have minimal impact, and others may even reduce the model's performance.

Two primary strategies are employed for fine-tuning:
\begin{itemize}
    \item \textbf{Hyper-Parameter Optimization:} This technique involves systematically adjusting the parameters of the augmentation functions to find a setting that lead to the best model performance, i.e., the highest improvement over the baseline performance. Techniques such as Bayesian optimization\cite{dewancker2016bayesian} or grid search~\cite{bergstra2011algorithms} are commonly used to explore the parameter space efficiently. For example, in a rain augmentation, parameters like raindrop size, density, and reflectivity might be adjusted to determine which combination yields the most improvement in the model's performance.

    \item \textbf{Latent Space Optimization:} This strategy leverages the internal feature representations of the \gls{dnn}. By extracting features from the intermediate layers of the \gls{dnn}, a classifier is trained to distinguish between clear weather features and features corresponding to specific \gls{odd} scenarios, such as fog. This classifier is then used to identify augmentations that produce feature representations similar to the challenging scenarios. The parameters of the identified augmentations are fine-tuned further, ensuring that they effectively simulate the conditions where the model previously struggled.
\end{itemize}

By fine-tuning the parameters of the augmentations, we ensure that the selected augmentations are not only relevant but also optimized to have the maximum possible impact on model robustness and performance.

\subsubsection{Augmented Training Process}
\label{sec:training-training-process}
With the augmentation functions selected and their parameters fine-tuned, the model undergoes training. 
The training process involves integrating the augmented data with the original training data. This integration is carefully managed to maintain a balance between the original and augmented images, ensuring that the model does not overfit to the augmented conditions while still improving its performance in those scenarios.

Several training strategies are explored to determine the most effective approach:
\begin{itemize}
    \item \textbf{Varying the Ratio of Augmented to Real Images:} Different ratios of augmented images to original images are tested to find the optimal balance. For example, training with a $1:1$ ratio of augmented to real images can be compared to a $1:2$ ratio, with performance evaluated on how well the model generalizes across all conditions.
    \item \textbf{Mini-Batching Augmentations:} This strategy involves grouping augmented images with their original counterparts in mini-batches during training. This approach can help the model learn to distinguish between subtle differences caused by augmentation and genuine environmental variations.
    \item \textbf{Tuning Augmentation Hyper-Parameters:} Beyond selecting the augmentations, the hyper-parameters controlling them are fine-tuned during training. For example, adjustments might be made to the extent of rotation or the level of noise added in certain augmentations.
\end{itemize}

Each of these strategies is tested experimentally to determine which provides the best improvement in the model’s robustness and overall performance.

\subsubsection{Robustness Evaluation}

The final stage of the methodology is the evaluation of the retrained model's robustness. This involves testing the model on both original and augmented datasets, particularly under the challenging conditions identified earlier. 
For instance, in the context of object detection under rainy conditions, the model's \gls{map} score is expected to improve significantly after training with rain augmentations. Additionally, other metrics such as the reduction in \gls{vr} and \gls{fr} are analyzed to confirm the effectiveness of the augmented training process.
A detailed discussion of these performance metrics can be found in \Cref{sec:eval-metrics}.

This thorough evaluation process not only confirms that the model’s performance has improved but also provides insights into the most effective augmented training strategies and augmentation techniques. By comparing the performance of models trained with different strategies, the methodology helps to identify the optimal training configuration for each specific use case.

For the specific datasets and models we use in our experiments, see \Cref{sec:experiment-setup}.

\subsection{Evaluation Metrics} \label{sec:eval-metrics}
To assess the performance of object detection and segmentation models, we utilize two evaluation metrics: \gls{map} and \gls{miou}. These metrics have become the de-facto standards in the field of computer vision, and we employ them to evaluate the accuracy and effectiveness of our models.

We use \gls{map} to evaluate the performance of our object detection models. By calculating the average precision of our models in detecting objects across multiple classes, we gain an understanding of their ability to accurately identify objects. We compute the precision of a detection model by taking the ratio of true positives to the sum of true positives and false positives, and then calculate the average precision by taking the mean of the precision values across all classes. This approach allows us to assess the overall performance of our models and identify areas for improvement.
The \gls{map} is calculated as
\begin{equation}
    \mathrm{AP} = \sum_{i=1}^{n-1}{(r_{i+1} - r_i)p_{\mathrm{interp}}(r_{i+1})},
\end{equation}

\begin{equation}
    \mathrm{\gls{map}} = \frac{\sum_{i=1}^{K}{AP_i}}{K},
\end{equation}
where the $\mathrm{AP}$ is calculated for each class $k$ and the \gls{map} averages over all classes.

In contrast, we employ \gls{miou} to evaluate the performance of our segmentation models. By calculating the average \gls{iou} of our predicted segmentation maps with the ground truth maps, we directly assess the accuracy of our models in segmenting objects. We compute the \gls{iou} by taking the ratio of the intersection area to the union area of the predicted and ground truth masks, and then calculate the \gls{miou} by taking the mean of the \gls{iou} values across all classes. This approach enables us to evaluate the effectiveness of our models in segmenting objects and identify opportunities for refinement.

To further assess the performance of our object detection models, we also employ two additional metrics: \gls{fr} and \gls{vr}:
\begin{equation}
    \mathrm{\gls{vr}} = \frac{\text{False Negatives}}{\text{True Positives} + \text{False Negatives}},
\end{equation}
\begin{equation}
    \mathrm{\gls{fr}} = \frac{\text{False Positives}}{\text{True Positives} + \text{False Positives}}.
\end{equation}
These metrics provide valuable insights into the types of errors made by our models. The \gls{fr} measures the proportion of objects that are incorrectly predicted, or ``fabricated'', relative to the total number of predicted objects. A lower \gls{fr} value indicates that our model is less prone to generating false positives. Conversely, the \gls{vr} measures the proportion of objects that are missed, or ``vanish'', relative to the total number of ground-truth objects. A lower \gls{vr} value indicates that our model is more effective at detecting all relevant objects.

\subsection{Augmented Training Strategy Experiment Design}
\label{sec:experiment-design-setup}

So far, we have introduced our augmented training process, which fine-tunes DNNs with augmentations selected to improve the specific model's robustness. This augmented training process has several possible strategies for both augmentation selection (\Cref{sec:fine-tuning-augmentations}) and augmented training process (\Cref{sec:training-training-process}). One can therefore use optimization to select an optimal augmented training strategy; this comes at high computational cost. Instead, one can aim to understand the relationship between a specific augmented training strategy and \gls{ml} model robustness.  To that end, we design and run \gls{ml} experiments.  
In this section, we present a general template we use to design the experiments presented in \Cref{sec:experimental-results}.

Designing an \gls{ml} experiment for testing the efficacy of an augmented training strategy requires one to specify:
    \begin{enumerate}
        \item A training paradigm, which includes (i) a source and target domain, (ii) a train / test / validation split, and (iii) training configurations, such as full training or fine-tuning. It can also include a strategy for hyper-parameter tuning, e.g., the \gls{ml} model’s hyper-parameters or the augmentation parameters (see \Cref{sec:fine-tuning-augmentations}).
        \item Augmentation selection, including a set of sensible parameter ranges within the parameterizable augmentations.
        \item Task-specific and general metrics that can be used to evaluate the augmented training experiment.
    \end{enumerate}

\begin{figure}[h]
\centering
\includegraphics[width=10cm]{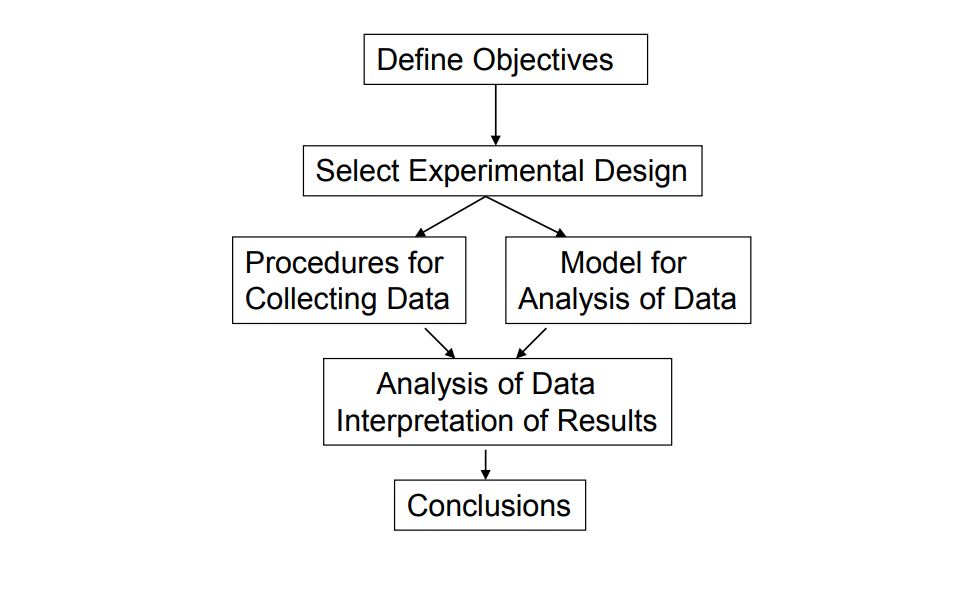}
\caption{General flowchart for designing an experiment from \cite{lawson2014design}.}
\label{fig:experiment-design}
\end{figure}

Our design for an augmented training experiment is based on the following framework visualized in \Cref{fig:experiment-design}, inspired by \cite{lawson2014design}. 

An augmented training strategy experiment's objective can be to either test a hypothesized training method, or to explore the influence of some training method's parameters on the model's performance. 
The next step, selecting an experiment design, is defining a procedure for collecting data (in this case, generating augmentations) and selecting a model (in this case, a statistical test, metric and test set) for analyzing the data. The final steps result from previous choices. In the following section, we expand each step in \Cref{fig:experiment-design} and illustrate each step with some examples.

\subsubsection{Experiment Planning Template}

\label{sec:experiment-planning-template}
To design an augmented training strategy experiment, we go through the following steps:

\paragraph{Step 1: Define the objectives of the experiment.} To do this, we can think through the following questions: Is the purpose of this experiment to (i) identify sources of variability, or to (ii) study cause-and-effect relationships? If the answer is (i), then is the goal to perform a screening experiment (i.e., screen for many potential variables), or is it an optimization experiment (i.e., the goal is to quantify the effects of a few factors)? 

For example, we could have the objective to determine how the augmented training parameters listed in \Cref{sec:training-training-process} impact the robustness of a model. This would be a screening experiment. Alternatively, we can decide to optimize one or more strategies, such as determining the best ratio of augmented to real image to use in training. If we choose to study a cause-and-effect relationships between an augmented training strategy and a model's robustness, we then need to decide on a meaningful observed effect size. Once the objective is defined, we can decide what remains fixed in the experiment, and what factors vary.

\paragraph{Step 2: Define what is being measured, what varies, and what remains fixed.} Augmented training strategy experiments measure the performance of a model on a selected test set. Before running an experiment, we must define a meaningful metric to show performance change; for example, the percent change in \gls{map} of a retrained model relative to a baseline model. In an augmented training scheme, we have parameters that are fixed, such as the train / test / validation data split, and other parameters that vary, such as the specific training strategy. In this step, we list exactly \textit{what} parameters we plan to change in our experiment. In the experiment design step, we decide \textit{how} we plan to vary our parameters.  

\paragraph{Step 3: Describe one run of the experiment.} A run of an augmented training experiment is to train an \gls{ml} model with a single configuration of a training strategy. It is important to describe how to do this exactly, including:
\begin{itemize}
\setlength\itemsep{-0.75em}
    \item Describe the datasets used,
    \item Compute and record the train / test / validation splits,
    \item Decide the training hyperparamters, such as number of epochs and learning rate,
    \item Describe one configuration of an augmented training strategy.
\end{itemize}
This step is essential in determining reproducibility of the experiment.

\paragraph{Step 4: Choose the experimental design.} The \textit{experimental design} is is the collection of runs that we will perform, including how we will vary the configuration of an augmented training strategy. A common choice is to take the parameters we selected to vary in Step 2 and follow a design, such as a factorial design or a randomized block design (for more information, see \cite{lawson2014design}). 

For example, suppose we want to run an experiment where we vary a single training parameter: the ratio of augmented to real images. How will we vary this, e.g., will it take every value in the set $\{1:1, 1:2, 1:3\}$? Or will we also consider a larger range? Do computational constraints restrict how we can test this parameter? This step aims to answers these questions.

\paragraph{Step 5: Determine a cross-validation scheme.} How many repetitions of each run do we need to perform in order to obtain measurable, meaningful results? In \gls{ml} experiments, this is done via a cross-validation scheme. In this step, we need to decide (i) how many cross-validation folds do we need to use? and (ii) how do we assign data to each fold? In the case of augmented training, where we retrain with real and augmented data, question (ii) is especially important to answer.

\paragraph{Step 6: Decide on a data analysis scheme.} Returning to Step 1, our data analysis scheme will enable us to meet our objective. Will we perform a statistical test, which could show a cause-and-effect relationship? Would we perform parametric or non-parametric tests? Or, does it suffice to compare the cross-validation averages of different runs? If we choose a statistical test, then we return to Step 5 to ensure that we have enough repetition (cross-validation folds) to perform the test. If computational constraints limit the number of repetitions, then we could consider a non-parametric test designed for a small sample size.

\begin{figure}[h!]
\centering
\includegraphics[width=14cm, trim={0 15cm 5cm 0},clip]{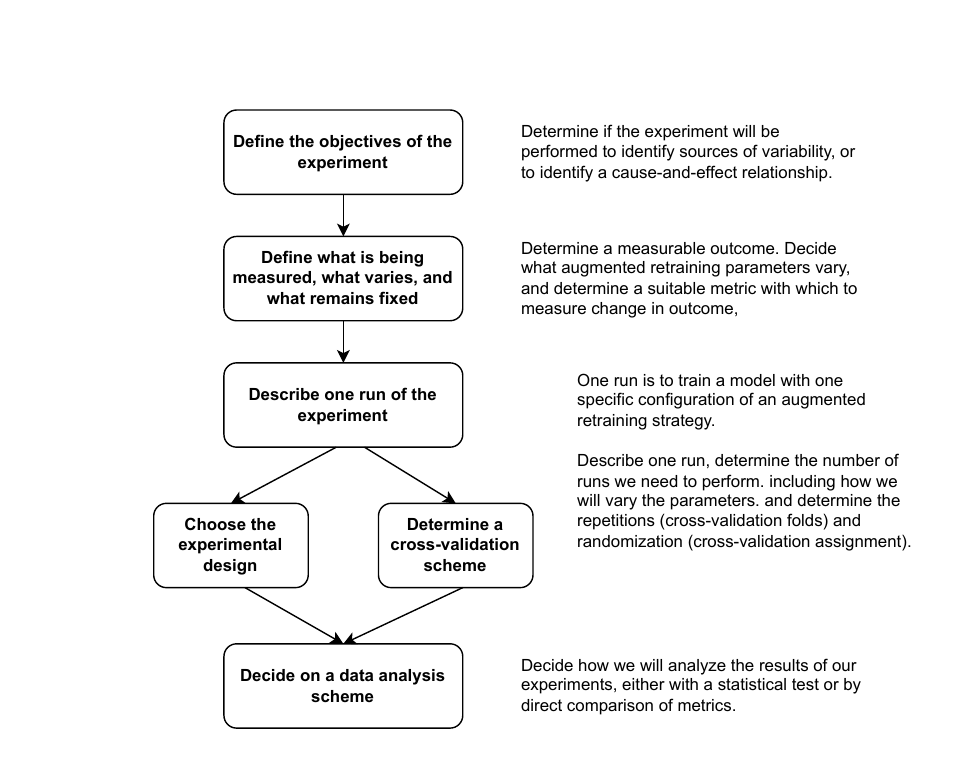}
\caption{Visualization of the template for designing an augmented training experiment.}
\label{fig:experiment-design-neurocat}
\end{figure}

\Cref{fig:experiment-design} illustrates the six steps of our previously introduced experiment planning template.
We observe that Steps 1, 2, and 3 follow from each other and determine how we will run the experiment.
Once we have the formal language to describe an experiment, we can choose how to evaluate it.
For this, we perform Step 4 and 5.
Finally, we perform Step 6 to decide how to analyze the results of the training strategy experiment.

As we will see in \Cref{sec:experimental-results}, finding a training strategy which incorporates augmented images is not as trivial as we might think.
Since there are numerous variables we can change for each training experiment, we must keep track and follow our experiment planning template to ensure that we find a reproducible and reliable training strategy.

\section{Experiment Setup}\label{sec:experiment-setup}

In \Cref{sec:overview-augmented-training-strategy} we presented our strategy for fine-tuning an \gls{ml} model with customized augmentations. We then described an experiment design for testing the impact of an augmented training strategy on model performance in \Cref{sec:experiment-design-setup}. In this section, we set up an experiment with the goal to identify the impact of using an augmented training strategy  to improve various \gls{ml} models' robustness to rainy-weather data. Using the template visualized in \Cref{fig:experiment-design-neurocat}, we give a high-level overview of our experiment, with details in the following sections. 

The objective of our experiment is to answer the question: \textit{``Does fine-tuning an \gls{ml} model with rainy-weather augmentations improve the model's robustness to rainy-weather conditions?''}. 
The augmented training strategy selected was we fine-tune a model with real and augmented images. 
We vary the ratio of augmented to real images during the fine tuning. 
Model robustness to rainy-weather conditions is measured by the model's  performance on a held-out real rainy-weather data relative to a baseline model.

In the following, we present details about the training paradigm of our experiments (see Section \ref{sec:experiment-design-setup}), which includes datasets, data splits, training configurations, and models.

\subsection{Datasets}
We use open-source datasets to train the models, generate augmented images, and test the performance of the trained model. The datasets are selected based on how well they satisfy the following conditions:
\begin{enumerate}
\setlength\itemsep{-0.75em}
    \item Relevance to autonomous driving,
    \item Quality and quantity of images,
    \item Accuracy and quality of ground truth annotations,
    \item Availability of images in adverse weather conditions.
\end{enumerate}

We test our approach on two use cases associated with autonomous driving: object detection and semantic segmentation.

\begin{figure}[h]

\begin{subfigure}{.475\linewidth}
  \includegraphics[width=\linewidth]{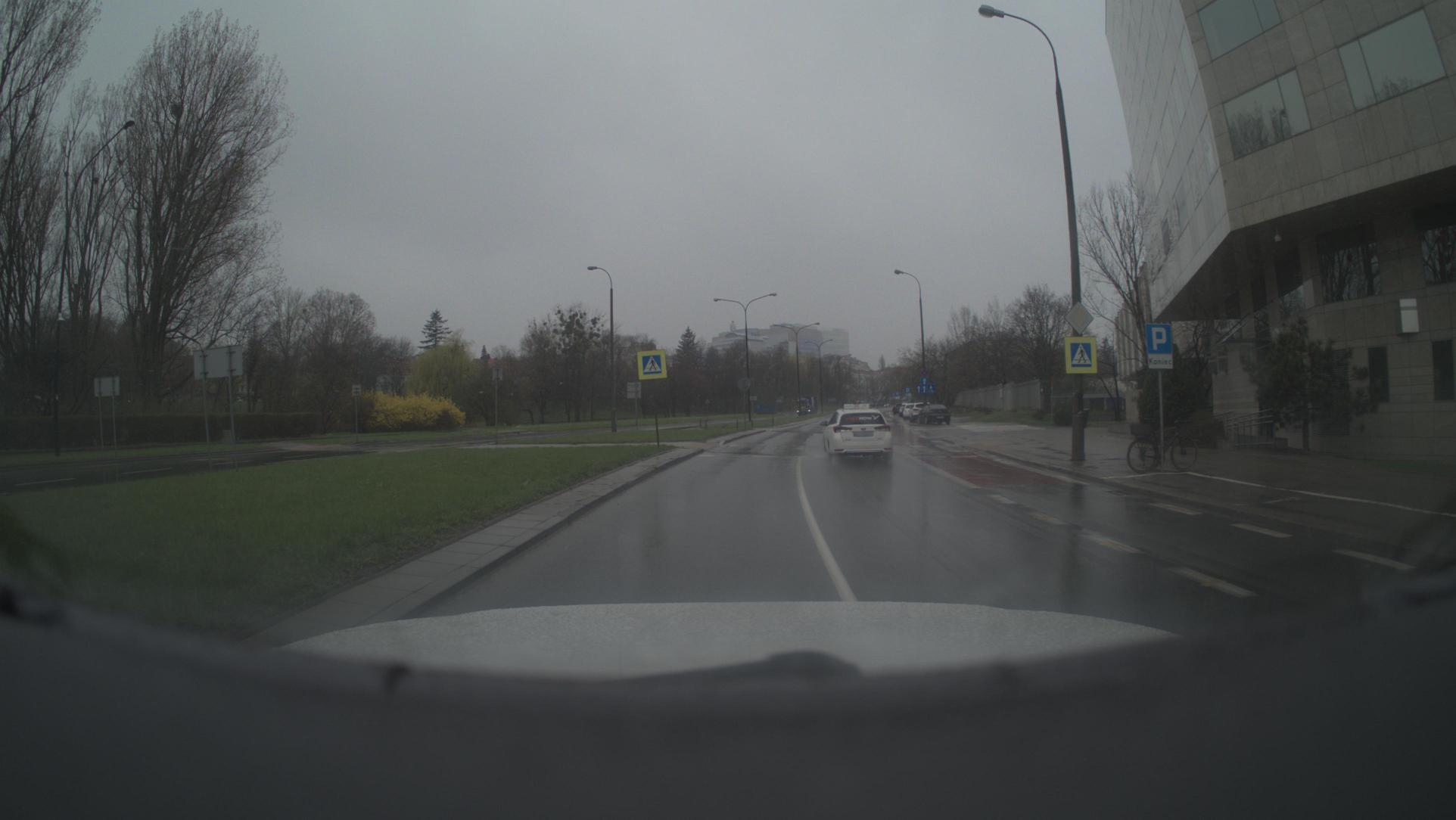}
  \caption{}
  \label{fig:rain_1}
\end{subfigure}\hfill 
\begin{subfigure}{.475\linewidth}
  \includegraphics[width=\linewidth]{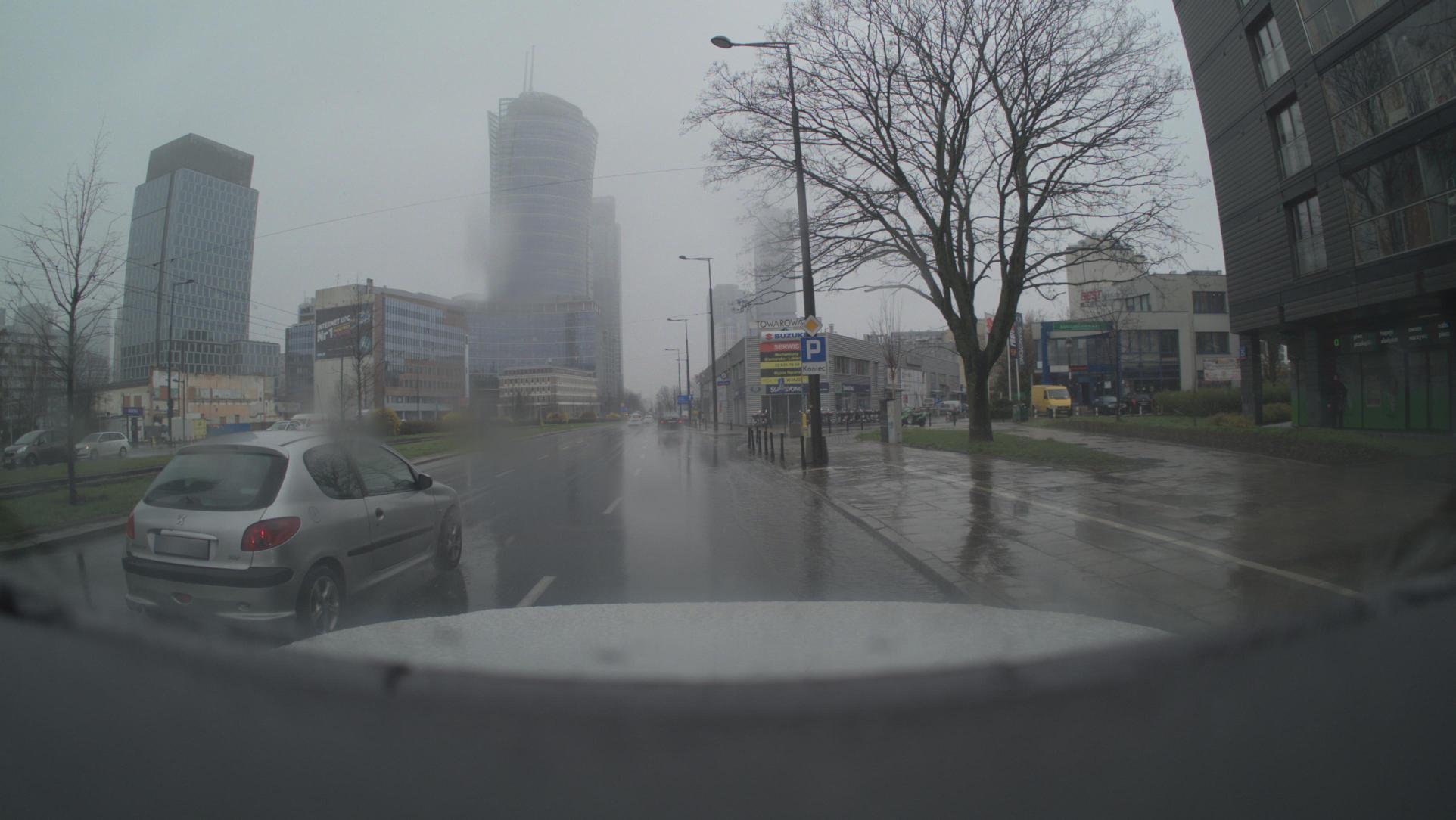}
  \caption{}
  \label{fig:rain_2}
\end{subfigure}

\medskip 
\begin{subfigure}{.475\linewidth}
  \includegraphics[width=\linewidth]{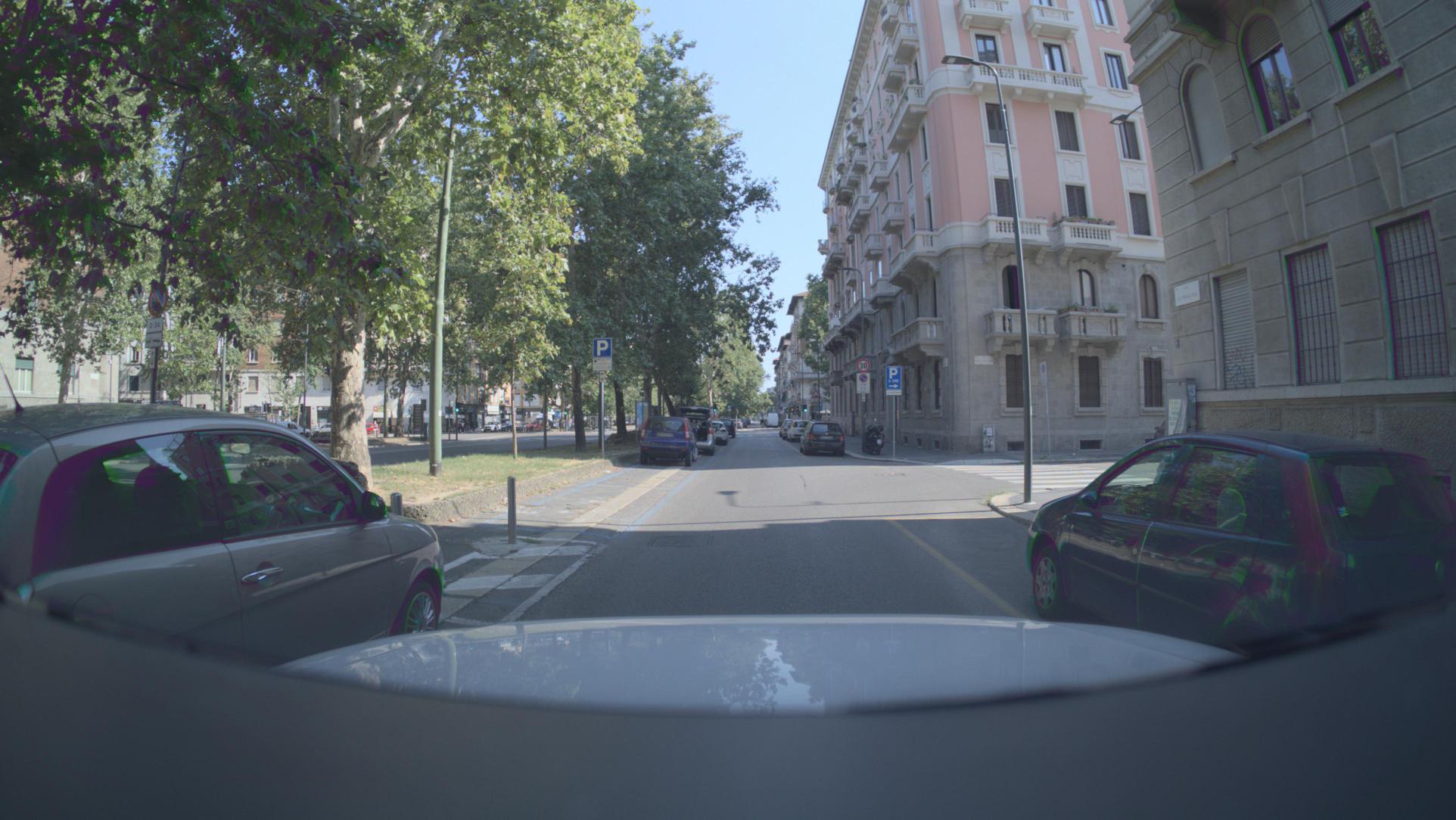}
  \caption{}
  \label{fig:clean_1}
\end{subfigure}\hfill 
\begin{subfigure}{.475\linewidth}
  \includegraphics[width=\linewidth]{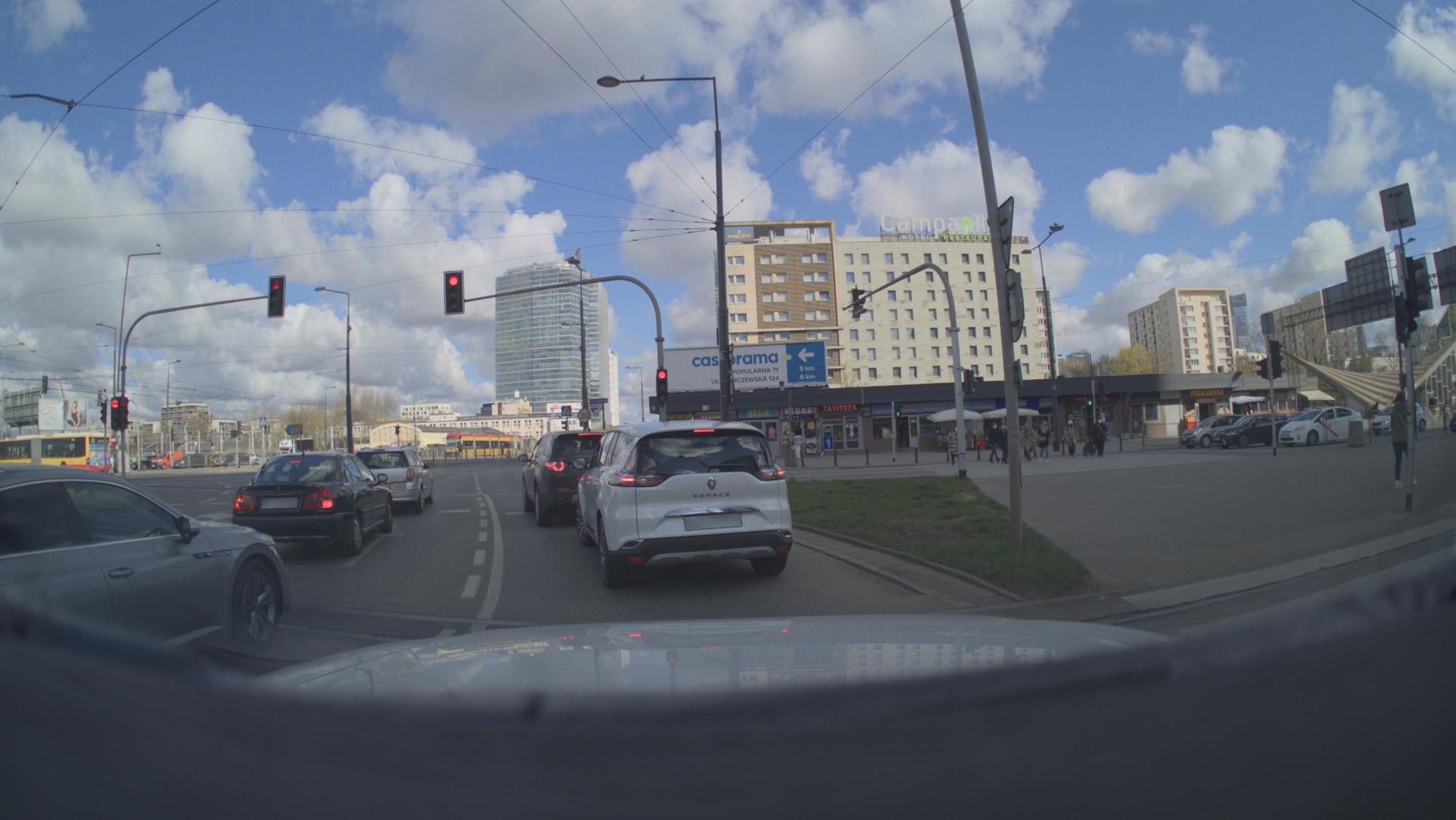}
  \caption{}
  \label{fig: clean_2}
\end{subfigure}

\caption{Example images from \gls{zod} dataset~\cite{zod_dataset}. (a), (b) are images collected during rain, while (c), (d) were collected during clear weather conditions.}
\label{fig:zod_example}
\end{figure}
\subsubsection{Dataset for Object Detection} 
\textbf{\gls{zod}}

The \gls{zod} \cite{zod_dataset} contains high-quality images and sequences for developing and testing perception models for autonomous driving systems. The $100,000$ images, collected in ideal and adverse weather conditions like clear skies, rain, and snow, as well as accurate ground truth labels, make it a popular choice for developing robust object detection models. It also includes images from different driving environments, such as urban and highways, and contains images collected during various times of the day. It stands as a valuable resource for autonomous driving applications.

During our experimentation, we found that the \gls{zod} data tagged as ``rainy'' images did not contain visually similar images, e.g., in some images it had been raining and the ground was still wet whereas other images were taken in the middle of a downpour.
To this end, we decided to tag the specific taxonomy of the rain in some of the \gls{zod} images, e.g., the color of the clouds, whether rain drops were present on the windshield, or if the road was visibly wet or not.
Exploring this taxonomy goes beyond the scope of this technical report, however, it could play a role when we want to robustify an \gls{ml} model on a more specific \gls{odd}.
\subsubsection{Dataset for Semantic Segmentation }
\textbf{Cityscapes Dataset}

\begin{figure}[ht!]

\begin{subfigure}{.475\linewidth}
  \includegraphics[width=\linewidth]{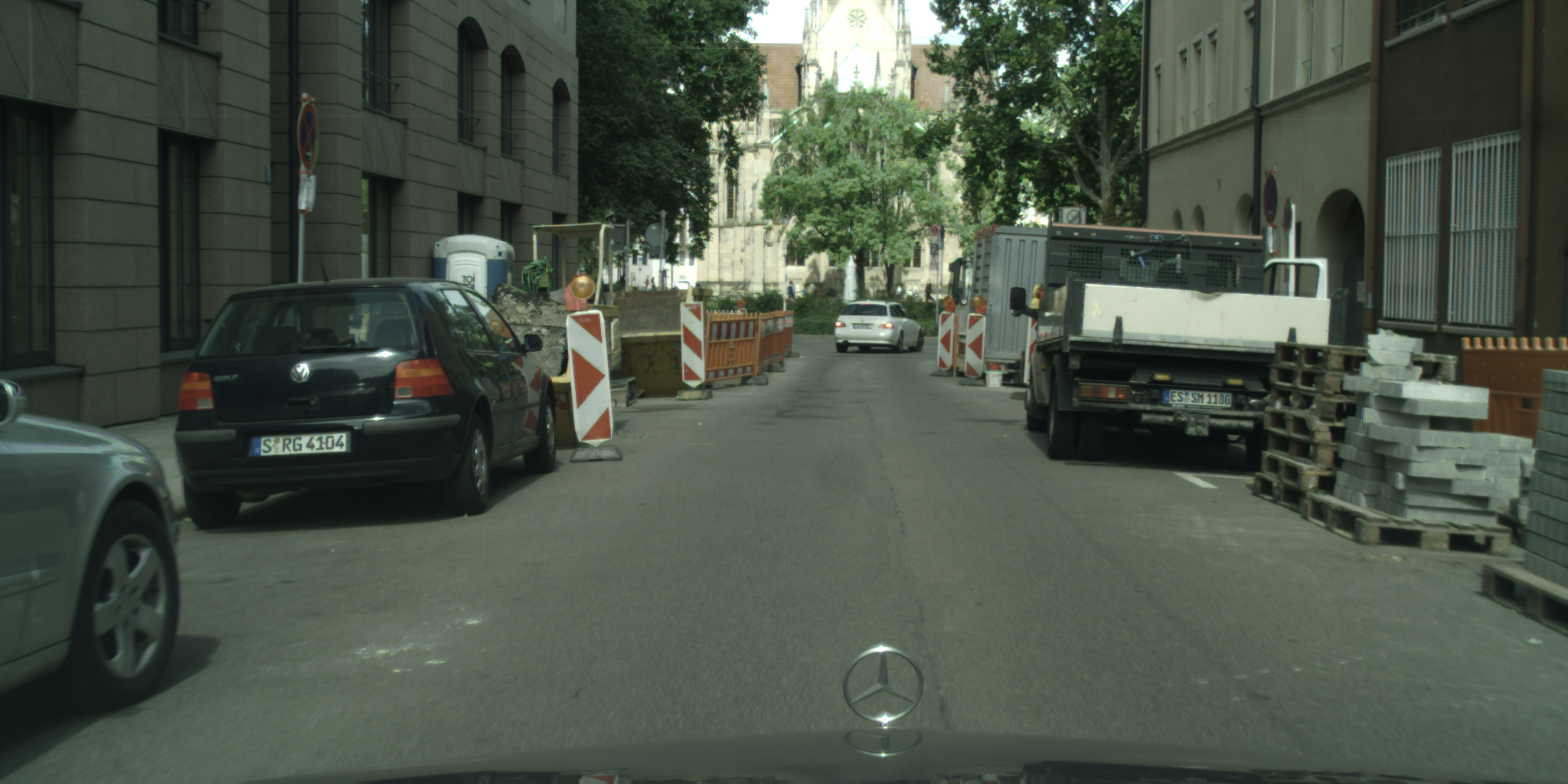}
  \caption{}
  \label{fig:city_1}
\end{subfigure}\hfill 
\begin{subfigure}{.475\linewidth}
  \includegraphics[width=\linewidth]{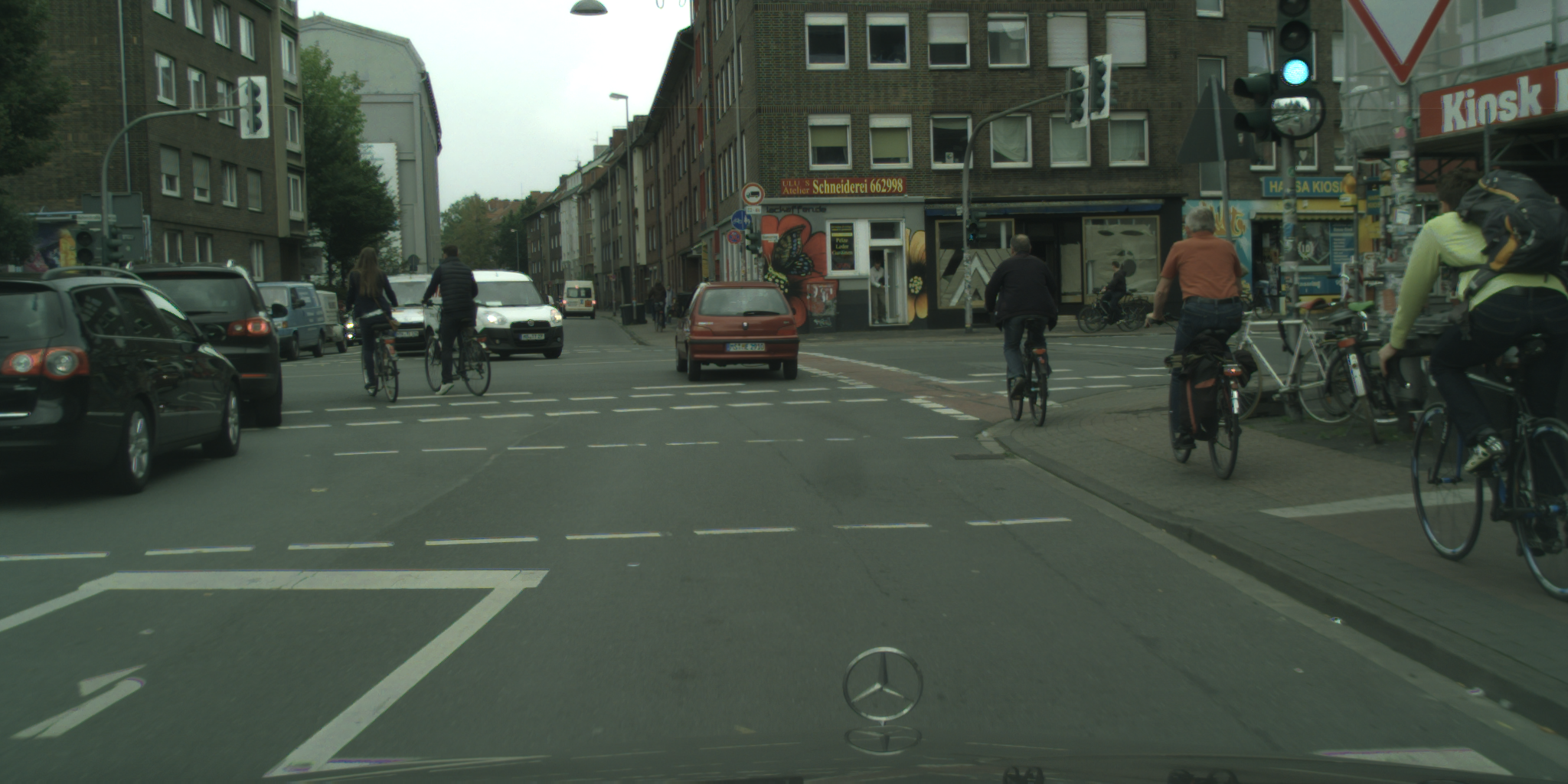}
  \caption{}
  \label{fig:city_2}
\end{subfigure}
\caption{Examples of clear weather images from the Cityscapes dataset~\cite{cityscapes_dataset}.}
\label{fig:city_example}
\end{figure}

The Cityscapes Dataset \cite{cityscapes_dataset} supports semantic segmentation tasks in urban street scenes. It features many high-resolution images collected from multiple cities, providing diverse urban environments focusing on scene understanding. While it primarily captures daytime scenes with good visibility, it also includes challenging lighting and weather conditions, such as overcast skies and wet roads.
However, it does not have images captured during intense weather conditions.
It is typically used to train semantic segmentation models in urban environments for autonomous driving. 

\textbf{\gls{acdc}}

\begin{figure}[hbt!]

\begin{subfigure}{.475\linewidth}
  \includegraphics[width=\linewidth]{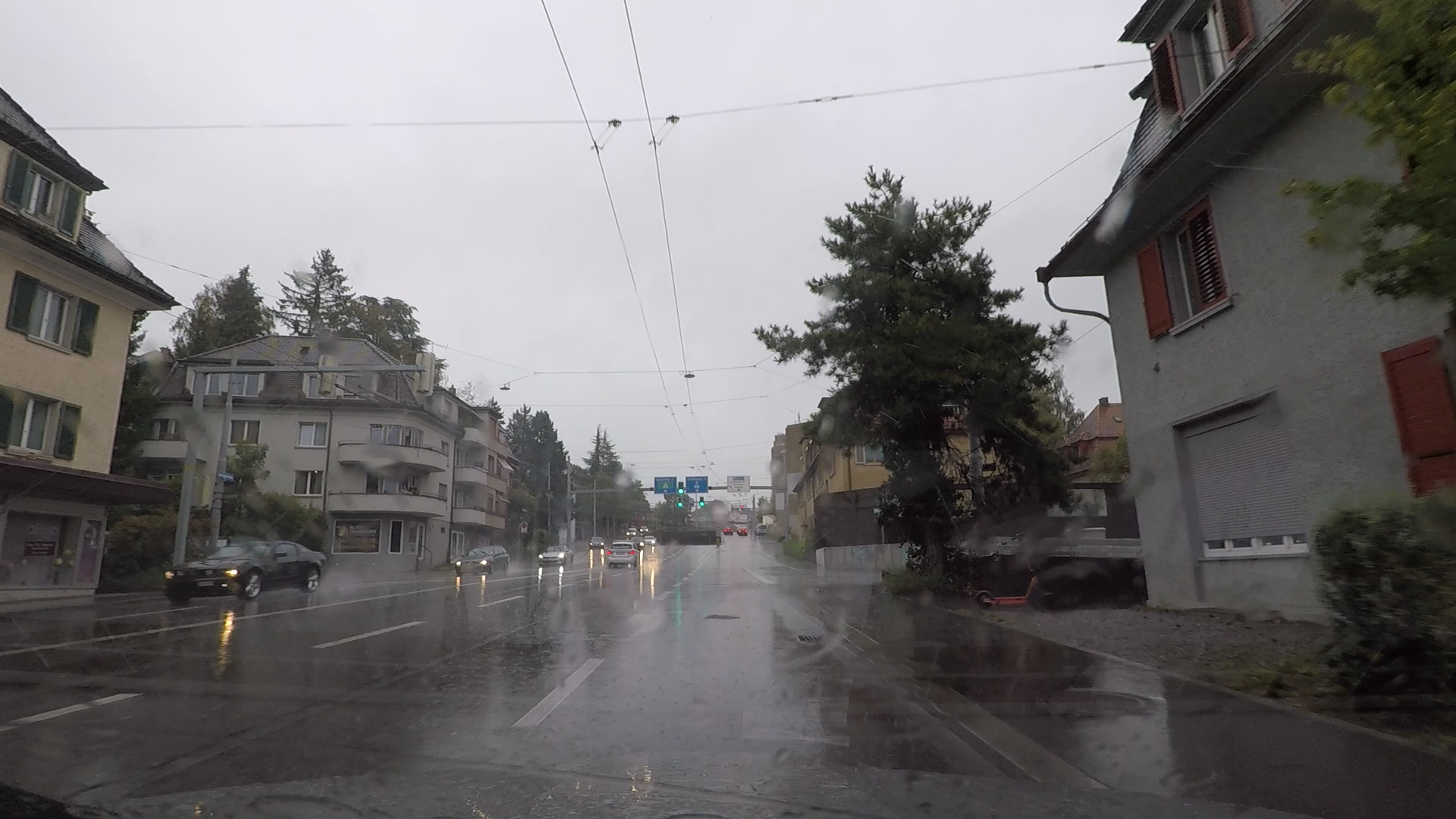}
  \caption{}
  \label{fig:acdc_rain_1}
\end{subfigure}\hfill 
\begin{subfigure}{.475\linewidth}
  \includegraphics[width=\linewidth]{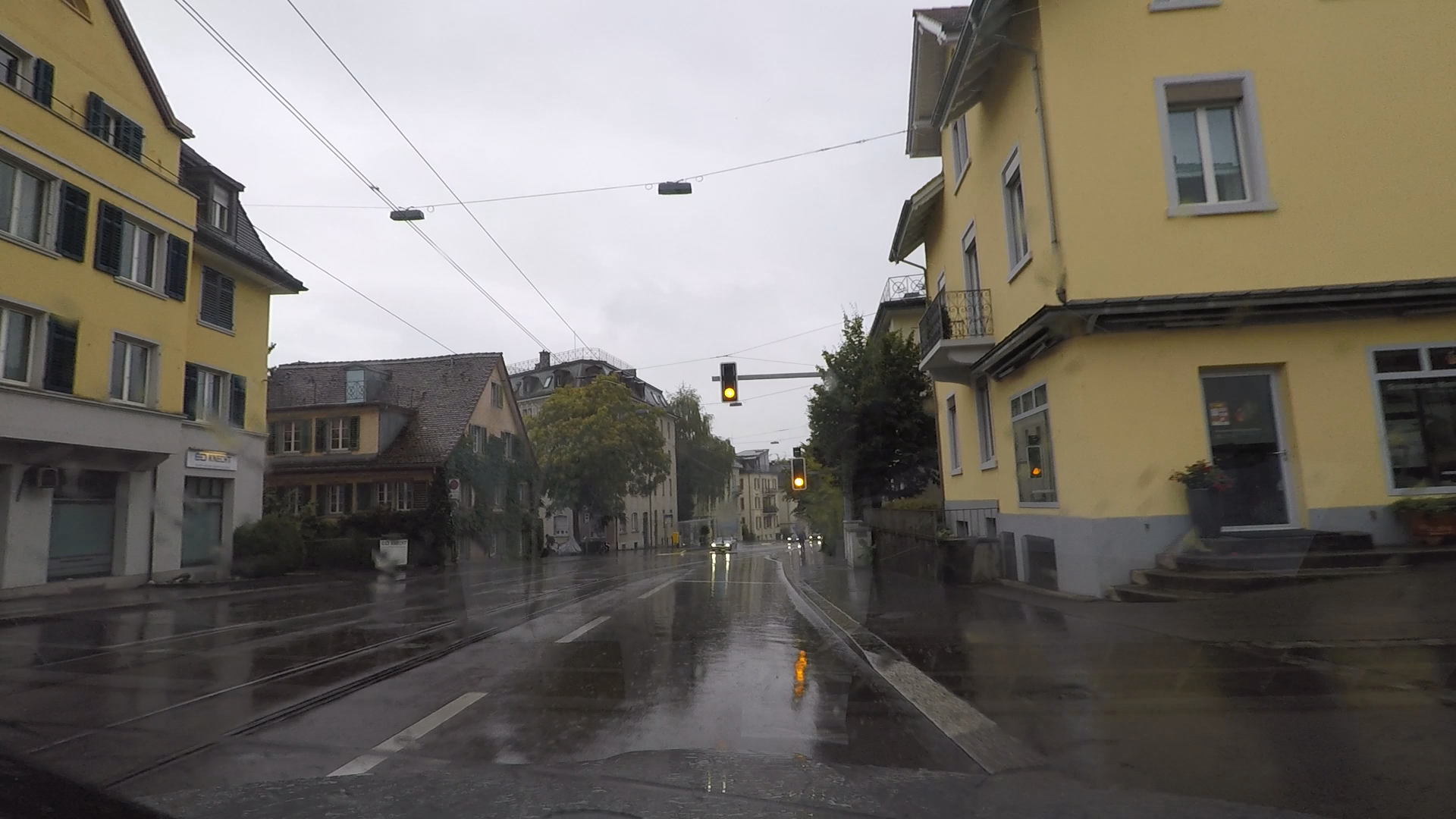}
  \caption{}
  \label{fig:acdc_rain_2}
\end{subfigure}
\caption{Examples of rain images from the \gls{acdc} dataset~\cite{acdc_dataset}.}
\label{fig:acdc_example}
\end{figure}

The \gls{acdc}~\cite{acdc_dataset}
provides high-quality images with pixel-level panoptic annotations for semantic perception tasks. It contains roughly $8,000$ images, nearly half of which are distributed across adverse visual conditions like rain, fog, snow, and night-time. \gls{acdc} helps researchers develop systems that perform consistently across various harsh weather environments.

\gls{acdc} was inspired by the Cityscapes dataset and aims to provide additional data for training and testing \gls{ml} models in adverse weather conditions. 
We chose it for our purposes since we aim to find a training strategy to improve the performance and robustness of semantic segmentation models on images containing adverse weather.
Moreover, since the data are collected in a similar manner to those from Cityscapes, the authors argue that the domain shift between the datasets should be minimal.

\subsection{Models}

We use pre-trained open source models for our experiments. This allows us to start from a solid baseline without having to spend additional time and resources. The models selected were also trained on open-source datasets and during our training process, we freeze certain layers in the backbone network to retain information from the pre-trained model. 

\subsubsection{Object Detection: Faster-RCNN}
This is an object detection model that uses the Faster R-CNN architecture and a ResNet50 model with a \gls{fpn} as backbone network for feature extraction \cite{frcnn}. The \gls{fpn} layer enhances the output features of the ResNet50 model and improves the detection of objects at different scales by constructing multi-scale feature maps. A \gls{rpn} generates anchor boxes for candidate regions which is then fine-tuned by a \gls{roi} layer. Finally separate regression and classification layers are used to identify the location of the object and its class, respectively.

\subsubsection{Semantic Segmentation: DeepLabV3+}

DeepLabV3+ \cite{deeplab} is a powerful semantic segmentation model that combines the standard encoder-decoder architecture with spatial pyramid pooling in the encoder and atrous convolutions in the decoder to capture finer semantic information. This combination improves object localization and is highly accurate in complex scenarios such as autonomous driving.

\subsection{Training Strategy and Evaluation}

Here, we describe one run of our experiment. For a selected model type, we begin with a pre-trained model and develop a baseline model by fine-tuning it using only clear weather images from its corresponding dataset. For example, we can create a baseline model from a model pre-trained on Cityscapes by fine-tuning it with a clear-weather Cityscapes subset. 

Then, the augmented training procedure done on the retrained model is as follows:
\begin{enumerate}
    \item  Select a ratio of real to augmented images, and generate a dataset with a combination of real and augmented images that matches this ratio. 
    \item Fine-tune the baseline model with the created dataset.
    \item Measure the model's performance on a held-out validation set.
\end{enumerate}

We repeat the above steps for different ratios of real to augmented images. Finally, we select the best-performing model to be evaluated on a held-out test set, along with the baseline model.

\subsubsection{Object Detection}\label{sec: obj_model_train}
We use the FRCNN-ResNet50 model with the \gls{zod} dataset for the object detection experiment. We start with a pre-trained model from PyTorch, where the ResNet50 backbone was trained on the ImageNet dataset~\cite{deng2009imagenet}, and the model layers were trained on the COCO dataset~\cite{lin2014microsoft}. The model was trained on $2,000$ clear weather images over six classes: vehicle, vulnerable vehicle, pedestrian, traffic sign, and traffic signal. An additional $1,000$ new clear weather images were used for validation and saving the best model based on the \gls{map} metric. The model was trained for $30$ epochs, after which there was no significant improvement in the best \gls{map} score. 

A dataset containing $1,000$ unseen clear weather images and $1,000$ images tagged as rain was selected to test the model. The \gls{map}, \gls{map}50, \gls{fr}, and \gls{vr} metrics were used to determine the performance of the best model obtained and set a baseline.

A custom rain augmentation generator was used to generate three unique augmentations for each image in the training set for the augmented training step. The pretraining model was trained on this new dataset so that the model and its augmentations are processed together for training at each step. Doing it this way led to more performance than randomly combining images and augmentations. The model was trained with the same parameters and configurations as the previous model, and the original's performance and the model trained on augmentations were compared.

\subsubsection{Semantic Segmentation}\label{sec: sem_model_train}
The DeepLabV3+ model uses a MobileNet backbone pre-trained on an ImageNet dataset~\cite{deng2009imagenet}. Like the object detection model, we train the base model on the clear weather dataset and test the model on both clear and adverse weather datasets. A problem we face in this case is that the Cityscapes dataset does not have annotations for the testing data or the adverse weather images. On the other hand, the \gls{acdc} dataset only has a limited number of images, which can lead to overfitting when used in training. To mitigate this issue, we trained the model on $2,075$ clear weather images of the Cityscapes and $300$ images from \gls{acdc}. The model was validated during training using $100$ Cityscapes images and $100$ \gls{acdc} images. Finally, the model was tested on $100$ Cityscapes clear weather images and $100$ \gls{acdc} rain images on the \gls{miou} metric.

\section{Experimental Results \& Discussion}\label{sec:experimental-results}
In this section, we describe the results we achieved when training various \gls{ml} models on object detection and semantic segmentation data introduced in the previous sections. We also indicate how we utilized the training strategies mention in \Cref{sec:methodology}.

\subsection{Object Detection}

We use the model trained only on clear weather images as defined in  \Cref{sec: obj_model_train} as the baseline for this experiment. This model is then tested on both unseen clear weather images and images with adverse rain weather. From Tables \ref{tab:det_clear} and \ref{tab:det_rain}, we can observe that the model has reduced performance on rain images when compared to clear weather images. This reduce in performance can be attributed to the fact that the model has never seen rain images before and thus is unable to generalize well on unseen datasets. 

We can now define our objective as improving the performance of the model on the rainy test set. To achieve this we start with a simple augmentation: horizontal flip. The idea behind this is introduce the model to new images and labels without actually collecting more data. This forces the model to generalize a bit more and thereby increase its performance on unseen data. We trained a model by flipping $2,000$ images and its labels from the dataset randomly. The results of this model show an improvement over the base model as seen in Tables \ref{tab:det_clear} and \ref{tab:det_rain} under the column \textbf{Model + Flip}. 

Next, we try to understand if this performance can be further improved by using the custom rain augmentations along with horizontal flipping. From our training dataset, we randomly flip $2,000$ images and also generated an augmented image for each of the $2,000$ images in order to improve our training dataset. From the column \textbf{Model + Flip + Aug}, we can see that the resultant model has a slightly better performance on certain metrics when compared to the model that was only trained on flipped images. A reason for this model to have lower performance on metrics like vanishing and fabrication ratio might be because of the augmentations being too out of distribution and thereby preventing the model from learning valuable information. 

To prevent the weights from drastically changing due to the introduction of augmentations, we apply the mini-batching training strategy as described in section \ref{sec:training-training-process}. For each image, we create $3$ unique augmentations and during a training step, the image along with its augmentations are processed as a batch by the model. From the column \textbf{Model + Batched Aug} in the tables \ref{tab:det_clear} and \ref{tab:det_rain}, we can see that this approach produced the best results across all the metrics for object detection. 

\begin{table}[h!]
\centering
\begin{tabular}{|c|c|c|c|c|}
\hline
\textbf{Metric} & \textbf{Baseline Model} & \textbf{Model + Flip} & \textbf{Model + Flip + Aug} & \textbf{Model + Batched Aug}\\
\hline
\gls{map} $\uparrow$ & 0.245 & 0.257& 0.259 &\textbf{0.263} \\
\hline
\gls{map}50 $\uparrow$ & 0.442 & 0.458 & 0.464 & \textbf{0.466} \\
\hline
\gls{fr} $\downarrow$ & 0.359 & 0.343 & {0.359} & \textbf{0.325} \\
\hline
\gls{vr} $\downarrow$& {0.443} & 0.433 & \textbf{0.429} & 0.438\\
\hline
\end{tabular}
\caption{Comparison between the model trained only on clear weather dataset and the models trained with custom rain augmentations. All of the models are tested on clear weather images. }
\label{tab:det_clear}
\end{table}

\begin{table}[h!]
\centering
\begin{tabular}{|c|c|c|c|c|}
\hline
\textbf{Metric} & \textbf{Baseline Model} & \textbf{Model + Flip} & \textbf{Model + Flip + Aug} & \textbf{Model + Batched Aug}\\
\hline
\gls{map} $\uparrow$  & 0.221 & 0.233 & 0.237 & \textbf{0.239} \\
\hline
\gls{map}50 $\uparrow$  & 0.414 & 0.431 & 0.435 & \textbf{0.441}\\
\hline
\gls{fr} $\downarrow$ & 0.372 & 0.351 & 0.373 & \textbf{0.345} \\
\hline
\gls{vr} $\downarrow$ & 0.450 & 0.439 & 0.434 &\textbf{0.432}\\
\hline
\end{tabular}
\caption{Comparison between the model trained only on clear weather dataset and the models trained with custom rain augmentations. The models are evaluated on 1,000 rain images from \gls{zod} dataset. 
}
\label{tab:det_rain}
\end{table}

\begin{figure}[h]
\centering
\includegraphics[width=15cm]{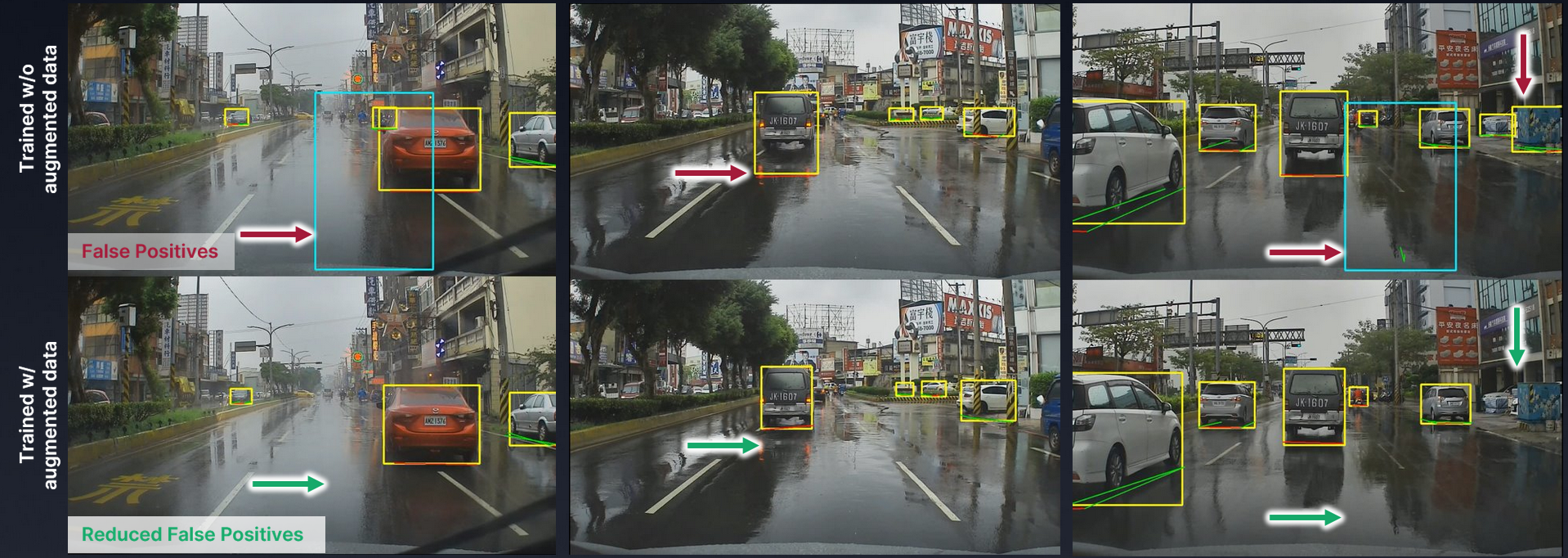}
\caption{The predictions of the models trained with and without augmentations on rainy images. The model trained with augmentations show lesser false positives than the model trained without augmentations.}
\label{fig: det_samples}
\end{figure}

\subsubsection{Discussion of Object Detection Results}
It is evident from the results in Tables \ref{tab:det_clear} and \ref{tab:det_rain} that we were able to satisfy our objective of improving the performance on the rainy test set without the need for collecting an additional dataset. Furthermore, we were also able to retain or even improve over the baseline performance on clear weather test set. Based on these results we have the following insights: 
\begin{itemize}
    \item The augmented training strategy is based on trial and error and each set of model and dataset requires different approaches to improve the performance of the model on out of distribution adverse conditions. 
    \item It is necessary to define our objectives beforehand so that we can monitor the improvement after each trial (see \Cref{sec:experiment-planning-template}). In \Cref{fig: det_samples}, we can see the effect the augmented training has on the false positive detections. If the initial objective was to decrease the number of false detections, then we could theoretically generate augmentations and identify training strategies that are more suited towards this goal. 
    \item The approach depends on domain knowledge and as a result the more we know about the initial model and the data distribution, the better the augmentations and the strategies used to obtain the desired result efficiently. 
    \end{itemize}

\subsection{Semantic Segmentation}
\begin{table}[h!]
\centering
\begin{tabular}{|c|c|c|c|}
\hline
\multirow{2}{*}{\textbf{Test set}} & \multicolumn{2}{|c|}{\textbf{mIoU score}} & 
\multirow{2}{*}{\textbf{Improvement [\%]}}\\
\cline{2-3}
& \textbf{Baseline Model} & \textbf{Model + Aug} & \\
\hline
\hline
Clear images & \textbf{0.7275} & 0.6480 & -10.93\\
\hline
Rain images & 0.3743 & \textbf{0.5356} & +43.09 \\
\hline
\end{tabular}
\caption{Comparison between the base model and the model trained with custom rain augmentations. The models are evaluated on both the clear weather images from Cityscapes and the rainy images from \gls{acdc}.}
\label{tab:seg_res}
\end{table}

While the results for object detection models showed an overall improvement on both test sets, we can observe that in the case of semantic segmentation, there is a significant trade-off between performance on clear-weather and adverse weather condition data. In Table \ref{tab:seg_res}, we see that on rainy images the retrained model achieves a 43.09\% mIoU improvement over the clear weather model. On clear weather images the performance of the retrained model drops by 10\% mIoU compared to the clear weather model. The trade-off requires further investigation on how to retain the improvement in rainy images without loosing the performance on clear weather images.

\subsubsection{Discussion of Semantic Segmentation Results}

Finding the training model presented in Table \ref{tab:seg_res}, required carefully modifying the model training used on clear weather images. In the following, we will present three major modifications we applied.

Firstly, we implemented a balanced validation set composed of both clear and adverse weather images. This approach was critical in preventing overfitting to either condition and provided a more accurate assessment of the model's generalization capabilities. Additionally, we introduced a weighted loss hyperparameter $\alpha$, which allowed us to control the influence of augmented images during training without altering the total number of training images. By setting the weight loss on clear-weather images to $1-\alpha$, we could influence the model's focus on different weather conditions while maintaining the consistency of other training parameters, such as the learning rate. Despite these efforts, hyperparameter tuning showed variability depending on the training data setup, suggesting that further experimentation with $\alpha$ might be necessary to optimize performance across all conditions.

Another factor influencing the results is the domain shift between the Cityscapes and \gls{acdc} datasets. Despite their similarities in camera setups and geographic recording locations, the differing environmental conditions may hinder the model's ability to fully leverage the augmented images. This domain shift underscores the importance of using clear and adverse weather images from the same dataset to minimize discrepancies and enhance model performance. The current reliance on adverse weather images primarily for testing limits our ability to establish an upper-bound of the performance, which could be accessed through a sufficiently large training set of adverse weather images.

In summary, while the augmented training process has shown promise in improving performance under adverse weather conditions, it also introduces challenges that require careful consideration of dataset composition, hyperparameter tuning, and the potential impact of a domain shift.

\section{Conclusion and Outlook} \label{sec:conclusion}

In this technical report, we discuss the \gls{ml} training strategy we developed to improve the performance of \glspl{dnn} on specific data domains.
Our training strategy begins by identifying the weak spots in an \gls{ml} model.	
We then create and customize augmentation functions to address these weak spots.
We introduce a detailed experiment design to help develop a training strategy to improve the model's performance on those weak spots.
Finally, we show how we implemented this training strategy to make an open-source object detection and semantic segmentation model more robust to specific weather conditions.

Our results and experiments showed us that finding a training strategy to improve performance is not trivial.
We found that varying different parts of the training strategy lead to different results.
Ultimately, we found a strategy which improved the relative performance of the object detection model by up to $+7.35\%$ in terms of \gls{map} on the clear weather images and $+8.14\%$ \gls{map} on rain weather data.
In terms of the semantic segmentation model, we saw an improvement of $+43.09\%$ on the rain images in terms of \gls{miou}.
These results are a promising step to help train models on custom augmentations.

We saw that training with complex augmentations benefits from additional \gls{ml} engineering.
We found that a training strategy including mini-batches containing both original and augmented versions of an input image led to larger performance improvements.
Optimizing the parameters of the augmentations can bring further improvements.
Ultimately, we found that finding an augmented training strategy is not a one-size-fits all approach.
Our experimental results show that the optimal variables for the training strategy are model and goal specific.

Due to time and budget limitations, we would have liked to continue experimenting with different variables in our training strategy to explore how much more we could improve the performance and robustness of \glspl{dnn} in automotive applications.
This is an important problem to solve since the more \gls{ml}-based driving functions are released, the more robust they should be to extensive \glspl{odd}.
In the future, we hope to verify our training strategy on other open-source models and datasets.
The insights gained from our research and the training strategies we developed, should be seen as a proof-of-concept that training on custom augmentations can improve the robustness of \glspl{dnn}.

\newpage
\printnoidxglossaries
\listoffigures
\newpage
\listoftables
\newpage


\bibliography{biblio}
\bibliographystyle{abbrv}

\end{document}